\documentclass{article}

\usepackage{microtype}
\usepackage{graphicx}
\usepackage{subfigure}
\usepackage{booktabs} % for professional tables

\usepackage{amsmath}
\usepackage{amssymb}
\usepackage{bm}
\usepackage{enumerate}
\usepackage{algorithmic}
\newcommand{\trace}{\mathop{\mathrm{Tr}}}

\usepackage{hyperref}

\usepackage[accepted]{icml2020}

\icmltitlerunning{Generative Flows with Matrix Exponential}

\begin{document}

\twocolumn[
\icmltitle{Generative Flows with Matrix Exponential}



\icmlsetsymbol{equal}{*}

\begin{icmlauthorlist}
\icmlauthor{Changyi Xiao}{USTC}
\icmlauthor{Ligang Liu}{USTC}
\end{icmlauthorlist}

\icmlaffiliation{USTC}{University of Science and Technology of China}

\icmlcorrespondingauthor{Ligang Liu}{lgliu@ustc.edu.cn}

\icmlkeywords{Machine Learning, ICML}

\vskip 0.3in
]



\printAffiliationsAndNotice{}  % leave blank if no need to mention equal contribution

\begin{abstract}
	Generative flows models enjoy the properties of tractable exact likelihood and efficient sampling, which are composed of a sequence of invertible functions. In this paper, we incorporate matrix exponential into generative flows. Matrix exponential is a map from matrices to invertible matrices, this property is suitable for generative flows. Based on matrix exponential, we propose matrix exponential coupling layers that are a general case of affine coupling layers and matrix exponential invertible $ 1\times 1 $ convolutions that do not collapse during training. And we modify the networks architecture to make training stable and significantly speed up the training process. Our experiments show that our model achieves great performance on density estimation amongst generative flows models. 
\end{abstract}

\section{Introduction}

Generative models aim to learn a probability distribution given data sampled from that distribution, in contrast with discriminative models, which do not require a large amount of annotations. A number of models have been proposed including generative adversarial networks (GANs) \cite{goodfellow2014generative}, variational autoencoders (VAEs)  \cite{kingma2013auto,rezende2014stochastic}, autoregressive models \cite{van2016pixel}, and generative flows models \cite{dinh2014nice,dinh2016density,rezende2015variational}. Generative flows models transform a simple probability distribution into a complex probability distribution through a sequence of invertible functions. They gain popularity recently due to exact density estimation and efficient sampling. In applications, they have been used for density estimation  \cite{dinh2016density}, data generation \cite{kingma2018glow} and reinforcement learning \cite{ward2019improving}.

How to design invertible functions is the core of generative flows models. There are two principles that should be followed. First, the Jacobian determinant should be computed efficiently. Second, the inverse function should be tractable. \citet{dinh2014nice} proposed coupling layers, they first applied generative flows models into density estimation. \citet{dinh2016density} extended the work with more expressive invertible functions and improved the architecture of generative flows models. \citet{kingma2018glow} proposed Glow: generative flow with invertible $ 1\times1 $ convolutions, which significantly improved the performance of generative flows models on density estimation and showed that generative flows models are capable of realistic synthesis. These flows all have easy Jacobian determinant and inverse.

However, generative flows models have not achieved the same performance on density estimation as state-of-the-art autoregressive models. In order to ensure that the function is invertible and effectively compute the Jacobian determinant, generative flows models suffer from two issues. First, due to the constraints of network, the network is not as expressive as that of GANs. Most of the flows constrain the Jacobian to a triangular matrix, which influences the effectiveness of the network. They apply element-wise transformation, parametrized by part of dimensions. \citet{dinh2014nice} proposed additive transformations, then \citet{dinh2016density} presented affine transformations. These transformations are very simple invertible transformations. In  \cite{huang2018neural,ho2019flow++}, they used invertible networks instead of affine transformations, but which still are univariate invertible transformations. Using univariate transformations also hurts the performance of generative flows models. Designing univariate invertible transformations is relatively simple compared with multivariate invertible transformations. Second, the dimension of latent space of generative flows models is same as the input space, which makes the network pretty large for high-dimensional data. \citet{dinh2016density} proposed a multiscale architecture to alleviate this problem, which gradually factors out a part of the total dimensions at regular intervals.

In this paper, we combine matrix exponential with generative flows to propose a new flow called matrix exponential flows. Matrix exponential can be seen as a map from matrices to invertible matrices, this property is suitable for constructing invertible transformations. Meanwhile, matrix exponential has other properties that are also helpful for generative flows. Based on matrix exponential, we propose matrix exponential coupling layers to enhance the expressiveness of networks, which can be seen as multivariate affine coupling layers. Training generative flows models often takes a long time to converge, especially for large-scale datasets. One reason is that training generative flows models is not stable, which prevents us from being able to use a larger learning rate. As standard convolutions may collapse during training, we propose a stable version of invertible $ 1\times 1 $ convolutions. And we also improve the coupling layers to make training stable and significantly accelerate the training process. The code for our model is available at \url{https://github.com/changyi7231/MEF}.

The main contributions of this paper are listed below:
\begin{enumerate}[1.]
	\item We incorporate matrix exponential into neural networks
	\item We propose matrix exponential coupling layers, which are a generalization of affine coupling layers.
	\item We propose matrix exponential invertible $ 1 \times 1 $ convolutions, which are more stable and efficient than standard convolutions. 
	\item We modify the networks architecture to make training stable and fast.
\end{enumerate}

%------------------------------------------------------------------------
\section{Background}
\subsection{Change of variables formula}
\label{section2.1}

Let $ X\in R^{n} $ be a random variable with an unknown probability density function $ p_{X}(x) $ and $ Z\in R^{n} $ be a random variable with a known and tractable probability density function $ p_{Z}(z) $, generative flows model is an unsupervised model for density estimation defined as an invertible function $ z=f(x) $ transforms  $X$ into $Z$. The relationship between $ p_{X}(x) $ and $ p_{Z}(z) $ follows\\
\begin{equation}
\log p_{X}(x)=\log p_{Z}(z)+ \log\left|\det(\frac{\partial f(x)}{\partial x})\right|
\label{formula:1}
\end{equation}
where $ \frac{\partial f(x)}{\partial x}$ is the Jacobian of $ f $ evaluated at $ x $. Note that a composition of invertible function remains invertible, let the invertible functions $ f $ be composed of $ K $ invertible functions: $ f_{K}\circ f_{K-1}\circ \cdots \circ f_{1}$. The log-likelihood of $ x $ can be written as\\
\begin{equation}
\log p_{X}(x)=\log p_{Z}(z)+\sum_{i=1}^{K}\log \left|\det(\frac{\partial h_{i}}{\partial h_{i-1}})\right|
\label{formula:2}
\end{equation}
where $ h_{i}=f_{i}\circ f_{i-1}\circ \cdots \circ f_{1}(h_{0}), h_{0}=x $\\
The choice of $ f $ should satisfy two conditions in order to be practical. First, computing the Jacobian determinant should be efficient. In general, the time complexity of computing Jacobian determinant is $ \mathcal{O}(n^3) $. Many works design different invertible functions such that the Jacobian determinant is tractable. Most of them constrain the Jacobian to a triangular matrix, which reduces the computation from $ \mathcal{O}(n^3) $ to $ \mathcal{O}(n) $. Second, in order to draw samples from $ p_{X}(x) $, the inverse function of $ f $ :$ x=f^{-1}(z) $ should be tractable. Since the generative process is the reverse process of inference process, generative flows models only use a single network. Generative flows models attain the capabilities of both efficient density estimation and sampling. 

\subsection{Generative Flows}
\label{section:2.2}

Generative flows models are constructed by a sequence of invertible functions, often parametrized by deep learning layers. Based on the two conditions mentioned in Section \ref{section2.1}, many generative flows have been proposed. We list several of them that are related to our model. See Table \ref{table:1} for an overview of these generative flows, and a description is as follows:

Affine coupling layers \cite{dinh2016density} partition the input into two parts. The first part of dimensions remains unchanged, and the second part of dimensions is mapped with an affine transformation, parametrized by the first part.

Since coupling layers only change half of dimensions, we need to shuffle the dimensions after every coupling layer. \citet{dinh2014nice} simply reversed the dimensions, \citet{dinh2016density} suggested randomly shuffling the dimensions. Since the operations are fixed during training, they may be limited in flexibility. \citet{kingma2018glow} generalized the shuffle operations to invertible $ 1\times 1 $ convolutions, which are more flexible and can be learned during training.

Actnorm layers \cite{kingma2018glow} are layers to improve training stability and performance. They perform an affine transformation of the activations using a scale and bias parameter per channel. They are data dependent, initialized such that the distribution of activations per-channel has zero mean and unit variance given an initial mini-batch of data.

\begin{table*}[t]
	\caption{The definition of several related generative flows and our generative flows. These flows all have easy Jacobian determinant and inverse. $ h, w, c $ denote the height, width and number of channels. The symbols $ \odot $, $ / $ denote element-wise multiplication and division. $ \boldsymbol{x},\boldsymbol{y} $ may denote the tensors with shape $ h \times w \times c $. 
	}
	\label{table:1}
	\vskip 0.15in
	\begin{center}
		\begin{small}
			\begin{tabular}{llll}
				\toprule
				Generative Flows & Function & Reverse Function & Log-determinant\\
				\hline
				Actnorm layers& $ \forall i,j: \bm{y}_{i,j,:}=\bm{s}\odot \bm{x}_{i,j,:}+\bm{b} $ & $ \forall i,j: \bm{x}_{i,j,:}=(\bm{y}_{i,j,:}-\bm{b})/\boldsymbol{s} $ &$ h\cdot w \cdot \mathrm{sum}(\log\left|\bm{s}\right|) $\\
				\hline
				Affine coupling & $[\bm{x}_{1},\bm{x}_{2}]=\bm{x}$ & $ [\bm{y}_{1},\bm{y}_{2}]=\bm{y} $ & $ \mathrm{sum}(\bm{s}(\bm{x}_{1})) $ \\
				layers & $ \bm{y}_{1}=\bm{x}_{1} $   & $ \bm{x}_{1}=\bm{y}_{1} $     & \\
				& $ \bm{y}_{2}=\exp(\bm{s}(\bm{x}_{1}))\odot \bm{x}_{2}+\bm{b}(\bm{x}_{1}) $ & $ \bm{x}_{2}=\exp(-\bm{s}(\bm{y}_{1}))\odot (\bm{y}_{2}-\bm{b}(\bm{y}_{1})) $ & \\
				& $ \bm{y}=[\bm{y}_{1},\bm{y}_{2}] $ & $ \bm{x}=[\bm{x}_{1},\bm{x}_{2}] $ & \\
				\hline
				Standard $ 1\times 1 $
				& $ \forall i,j: \bm{y}_{i,j,:}=\bm{W}\bm{x}_{i,j,:} $ &  $ \forall i,j: \bm{x}_{i,j,:}=\bm{W}^{-1}\bm{y}_{i,j,:} $ & $ h\cdot w \cdot \log\left|\det(\bm{W})\right| $\\
				convolutions & & &\\
				\hline
				Matrix exp & $[\bm{x}_{1},\bm{x}_{2}]=\bm{x}$ & $ [\bm{y}_{1},\bm{y}_{2}]=\bm{y} $ & $ \trace(\bm{s}(\bm{x}_{1})) $ \\
				coupling layers & $ \bm{y}_{1}=\bm{x}_{1} $   & $ \bm{x}_{1}=\bm{y}_{1} $     & \\
				See Section \ref{section:4.1} & $ \bm{y}_{2}=\bm{e}^{\bm{S}(\bm{x}_{1})} \bm{x}_{2}+\bm{b}(\bm{x}_{1}) $ & $ \bm{x}_{2}=\bm{e}^{-\bm{S}(\bm{y}_{1})} (\bm{y}_{2}-\bm{b}(\bm{y}_{1})) $ & \\
				& $ \bm{y}=[\bm{y}_{1},\bm{y}_{2}] $ & $ \bm{x}=[\bm{x}_{1},\bm{x}_{2}] $ & \\
				\hline
				Matrix exp $ 1\times 1 $ &  $ \forall i,j: \bm{y}_{i,j,:}=\bm{e}^{\bm{W}}\bm{x}_{i,j,:} $ &  $ \forall i,j: \bm{x}_{i,j,:}=\bm{e}^{-\bm{W}}\bm{y}_{i,j,:} $ & $ h\cdot w \cdot \trace(\bm{W}) $\\
				convolutions & & &\\
				See Section \ref{section:4.2} & & &\\ 
				\bottomrule
			\end{tabular}
		\end{small}
	\end{center}
	\vskip -0.1in
\end{table*}

%------------------------------------------------------------------------

\section{Matrix Exponential}

In generative flows models, the function $ f $ is implemented as a sequence of invertible functions, which can be parametrized by a neural network $ f=\bm{W}_{m}\phi(\bm{W}_{m-1}\phi(\bm{W}_{m-2}\cdots \phi(\bm{W}_{1}\bm{x}))) $, where $ \bm{W}_{i} $, $  1\leq i \leq m $, is the weight matrix and $ \phi $ is activation function. To ensure that $ f $ is invertible, we can let $ \bm{W}_{i} $ be an invertible matrix and $ \phi $ be a strictly monotone function. But it is difficult to ensure that $ \bm{W}_{i} $ is invertible during training, and computing the determinant of $ \bm{W} $ is $ \mathcal{O}(n^3) $ in general. One method is to enforce $ \bm{W}_{i} $
to be a triangular matrix. Its determinant is the product of its diagonal entries and it is invertible as long as its diagonal entries are non-zero. But triangular matrices are less expressive and computing the inverse of triangular matrices is sequential, not parallel. We propose to replace the weight matrix $ \bm{W}_{i} $ by the matrix exponential of $ \bm{W}_{i} $. It can make the networks invertible. And it is more expressive than triangular matrices. Moreover, computing the determinant needs only $ \mathcal{O}(n) $ time and computing the inverse matrix is easy and parallel. Before introducing matrix exponential, we first set up some notations. Let $ M_{n}(R)  $ be the set of $n \times n$ real matrices, $ GL_{n}(R)  $ be the set of $n \times n$  real invertible matrices, $ GL_{n}(R)^{+}  $ be the set of $n \times n$  real invertible matrices with positive determinant, $ GL_{n}(R)^{-}  $ be the set of $n \times n$  real invertible matrices with negative determinant.

\subsection{Properties of Matrix Exponential}
\label{section:3.1}

Matrix exponential is a matrix function whose definition is similar to the exponential function. Matrix exponential has many applications. It can be used to solve systems of linear differential equations. And it plays an important role in the theory of Lie groups, which gives the connection between a matrix Lie algebra and the corresponding Lie group.
Matrix exponential of $ \bm{W}\in M_{n}(R) $ is defined as
\begin{equation}
\bm{e}^{\bm{W}}=\sum_{i=0}^{\infty}\frac{\bm{W}^{i}}{i!}
\label{formula:3}
\end{equation}
Matrix exponential has four properties as follows:
\begin{enumerate}[1.]
	\item  For any matrix $ \bm{W}\in M_{n}(R) $, $ \bm{e}^{\bm{W}} $ is converge, and $ \bm{e}^{\bm{W}}\in GL_{n}(R) $, meanwhile $ (\bm{e}^{\bm{W}})^{-1}=\bm{e}^{-\bm{W}} $.
	\item  $ \log\det(\bm{e}^{\bm{W}})=\trace(\bm{W}) $.
	\item  For any matrix $ \bm{X}\in GL_{n}(R)^{+} $ and $ \bm{X} $ satisfies each Jordan block of $ \bm{X} $ corresponding to a negative eigenvalue occurs an even number of times, then there exists a matrix $ \bm{W}\in M_{n}(R) $ such that $ \bm{X}=\bm{e}^{\bm{W}} $.
	\item  For any matrix $ \bm{X}\in GL_{n}(R)^{+} $, there exists some matrices $ \bm{W}_{1}, \bm{W}_{2}, \dots, \bm{W}_{n}\in M_n(R) $ such that $ \bm{X}=\bm{e}^{\bm{W}_{1}}e^{\bm{W}_{2}}\cdots \bm{e}^{\bm{W}_{n}} $. 
\end{enumerate}
See \cite{hall2015lie} for the proofs of property 1,2,4 and \cite{culver1966existence} for the proof of property 3.  From property 1, we see that matrix exponential of $ \bm{W} $ is always converge and invertible. Thus we implement the neural network as $ f=\bm{e}^{\bm{W}_{m}}\phi(\bm{e}^{\bm{W}_{m-1}}\phi(\bm{e}^{\bm{W}_{m-2}}\cdots \phi(\bm{e}^{\bm{W}_{1}}\bm{x}))) $, which makes the neural network invertible. Matrix exponential can be seen as a map from $ M_{n}(R) $ to $ GL_{n}(R) $, so we have no need to constrain the weight matrix $ \bm{W} $. The inverse of matrix exponential is also matrix exponential, which can be computed in the same way. From property 2, computing the log-determinant of matrix exponential of $ \bm{W} $ turns into computing the trace of $ \bm{W} $. Computing the determinant of a $ n \times n $ matrix is $ \mathcal{O}(n^3) $ in general, while computing the trace is $ \mathcal{O}(n) $. Property 3 demonstrates the image of matrix exponential. Matrix exponential is not a surjective map from $ M_{n}(R) $ to $ GL_{n}(R) $. Property 2 shows that the determinant of matrix exponential is always positive. $ GL_{n}(R) $ has two connected components: $ GL_{n}(R)^{+} $ and $ GL_{n}(R)^{-}$. The image of matrix exponential is a subset of $ GL_{n}(R)^{+} $. It is reasonable to ensure that the determinant is positive, because once the sign of the determinant changes during training, it may cause the matrix to be singular and numerically unstable. Although the matrix exponential is also not a surjective map to $ GL_{n}(R)^{+} $, however, it only excludes a few matrices. Property 4 demonstrates any matrix in $ GL_{n}(R)^{+} $ is a product of $ n $ matrix exponentials, so we can get a surjective map to $ GL_{n}(R)^{+} $. In practice, using $ n $ matrix exponentials may be redundant, since the image of matrix exponential is a rich class of invertible matrices, one or two is enough. 

\subsection{Compute Matrix Exponential}
\label{section:3.2}

Matrix exponential is an infinite matrix series. Dozens of methods for computing matrix exponential have been proposed. \citet{moler2003nineteen} showed nineteen ways involving approximation theory, differential equations, the matrix eigenvalues. We propose two methods to incorporate matrix exponential into neural networks. The first method is for low-dimensional data, and the second method is for high-dimensional data. 

The first method is to truncate the matrix series of Eq. (\ref{formula:3}) at index $ k $ to approximate the matrix series. Define the finite matrix series as
\begin{equation}
T_{k}(\bm{W})=\sum_{i=0}^{k}\frac{\bm{W}^{i}}{i!}
\label{formula:4}
\end{equation}
There are several papers that study the truncation error of this series, \citet{liou1966novel} gave a bound of truncation error
\begin{equation}
\|T_{k}(\bm{W})-\bm{e}^{\bm{W}}\|_{1}\leq(\frac{\|\bm{W}\|_{1}^{k+1}}{(k+1)!})(\frac{1}{1-\|\bm{W}\|_{1}/(k+2)})
\label{formula:5}
\end{equation}
where $ \|\cdot\|_{1} $ is the matrix 1-norm. The error bound is affected by $ \|\bm{W}\|_{1} $, which decreases as $ \|\bm{W}\|_{1} $ decrease. When $ \|\bm{W}\|_{1} $ is small, we can choose a small $ k $ and need less computation to approximate the infinite series. Fortunately, the value of weight matrices of neural networks is often small such that $ \|\bm{W}\|_{1} $ is also small. This is a good property that makes incorporating matrix exponential into neural networks practical. Since $ \bm{e}^{\bm{W}}=(\bm{e}^{\bm{W}/2^s})^{2^s} $, we first scale the weight matrix to a smaller value, then compute the matrix exponential and the matrix power. This further reduces the computation. Algorithm \ref{alg:1} shows the process of computing matrix exponential, which is mentioned in \citet{moler2003nineteen}. This algorithm costs about $ (s+k-1)n^{3} $ FLOPs, which makes it unable to scale to high-dimensional data.
% But approximating the matrix exponential needs several times of matrix multiplication, which makes this algorithm unable to scale to high-dimensional data.

\begin{algorithm}[t]
	\caption{Algorithm for computing matrix exponential} 
	\label{alg:1}
	\begin{algorithmic}[1]
		\REQUIRE ~~\\ 
		Weight matrix: $ \bm{W} $\\
		%Scale coefficient: $ s $\\
		Tolerable  error: $ \epsilon $
		\ENSURE ~~\\ 
		Matrix exponential of weight matrix: $ \bm{e}^{\bm{W}} $
		\STATE choose the smallest non-negative integer s such that $ \|\bm{W}\|_{1}/(2^s)<\frac{1}{2} $
		\STATE $ \bm{W}:=\bm{W}/(2^s) $
		\STATE $ \bm{X}:=\bm{I} $
		\STATE $ \bm{Y}:=\bm{W} $
		\STATE $ k:=2 $
		\WHILE {$ \|\bm{Y}\|_{1}>\epsilon $} 
		\STATE $ \bm{X}:=\bm{X}+\bm{Y} $
		\STATE $ \bm{Y}:=\bm{W}\cdot \bm{Y}/k $
		\STATE $ k:=k+1 $
		\ENDWHILE
		\FOR {$ i=1 $ to $ s $}
		\STATE $ \bm{Y}:=\bm{Y}\cdot \bm{Y} $
		\ENDFOR
		\STATE \textbf{return:} $\bm{Y}$ 
	\end{algorithmic}
\end{algorithm}

We propose the second method that combines matrix exponential with neural networks for high-dimensional data. Instead of directly parameterizing the weight matrix $ \bm{W} $, we propose a low-rank parameterization method. Let $ \bm{W}=\bm{A}_{1}\bm{A}_{2}$, where $ \bm{A}_1\in R^{n\times t},\bm{A}_{2} \in R^{t\times n}$, and $\bm{A}_{1}, \bm{A}_{2} $ are the weight matrices. Substitute $ \bm{W} $ into Eq. (\ref{formula:3}), then
\begin{equation}
\bm{e}^{\bm{W}}=\sum_{i=0}^{\infty}\frac{(\bm{A}_{1}\bm{A}_{2})^{i}}{i!}
\label{formula:6}
\end{equation}
Let $ \bm{V}=\bm{A}_{2}\bm{A}_{1} $. Considering the associative law of matrix multiplication, we have
\begin{equation}
\bm{e}^{\bm{W}}=\bm{I}+\bm{A}_{1}\sum_{i=0}^{\infty}\frac{\bm{V}^{i}}{(i+1)!}\bm{A}_{2}
\label{formula:7}
\end{equation}
We truncate matrix series of $ \bm{V} $ at index $ k $ to approximate $ \bm{e}^{\bm{W}} $. Similar to the truncation error bound of $ \bm{e}^{\bm{W}} $, the error bound of truncating matrix series of $ \bm{V} $ is given by
\begin{equation}
\|\sum_{i=k+1}^{\infty}\frac{\bm{V}^{i}}{(i+1)!}\|_{1}\leq(\frac{\|\bm{V}\|_{1}^{k+1}}{(k+2)!})(\frac{1}{1-\|\bm{V}\|_{1}/(k+3)})
\label{formula:8}
\end{equation}
Computing the matrix series of $ \bm{W} $ turns into computing the matrix series of $ \bm{V} $. Since $ \bm{V}\in R^{t\times t} $, computing the matrix series of $ \bm{V} $ costs $ \mathcal{O}(t^3) $. 
The rank of matrix $ \bm{W} $ is less than or equal to $ t $. We can choose a small $ t $ to reduce the computation, but which will hurt the expressiveness. It is a balance between expressiveness and computation. Computing the matrix series of $ \bm{V} $ is analogous to Algorithm \ref{alg:1}, just set the scale coefficient $ s:=0 $ and modify the line 4 to $ \bm{Y}:=\bm{W}/2 $ and line 5 to $ k:=3 $.

%------------------------------------------------------------------------

\begin{figure}[t]
	\vskip 0.2in
	\begin{center}
		\subfigure[One step of flow]{
			\begin{minipage}[t]{0.4\linewidth}
				\includegraphics[width=1.0\linewidth]{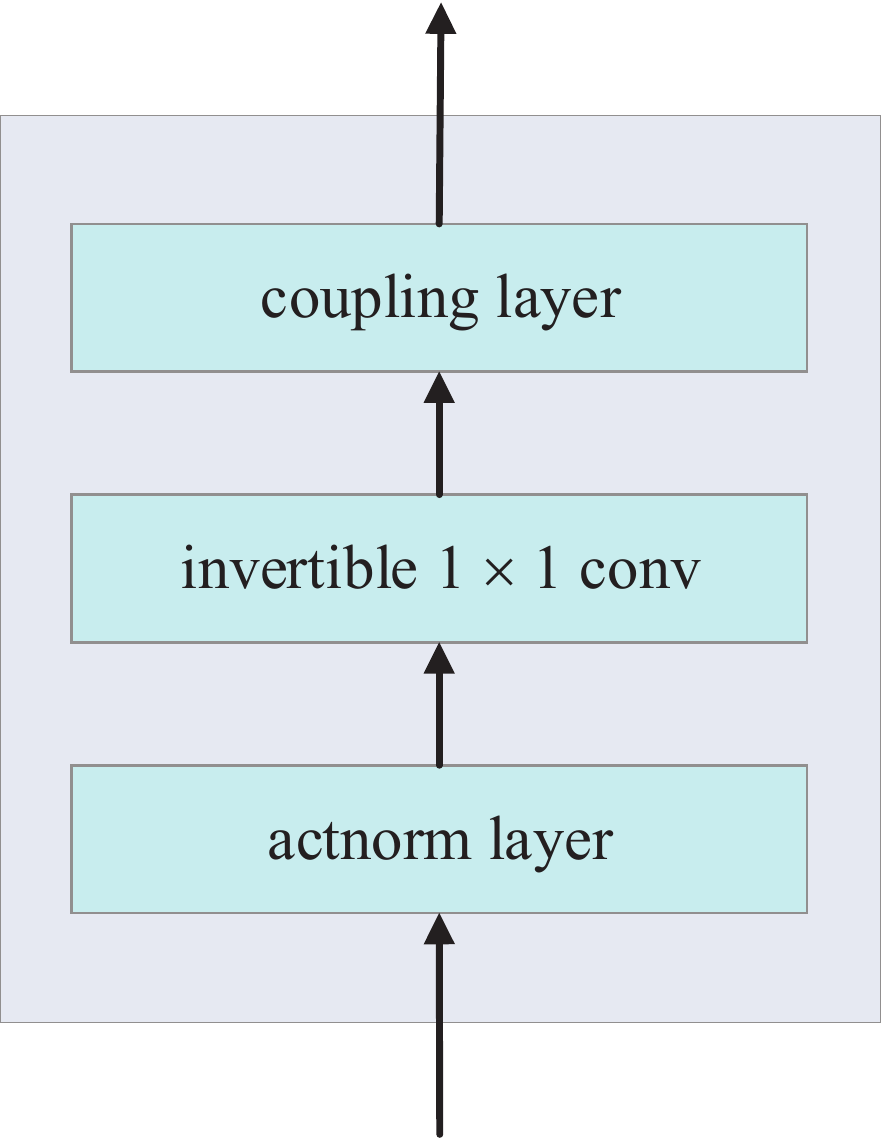}
		\end{minipage}}
		\subfigure[Model architecture]{
			\begin{minipage}[t]{0.45\linewidth}
				\includegraphics[width=1.2\linewidth]{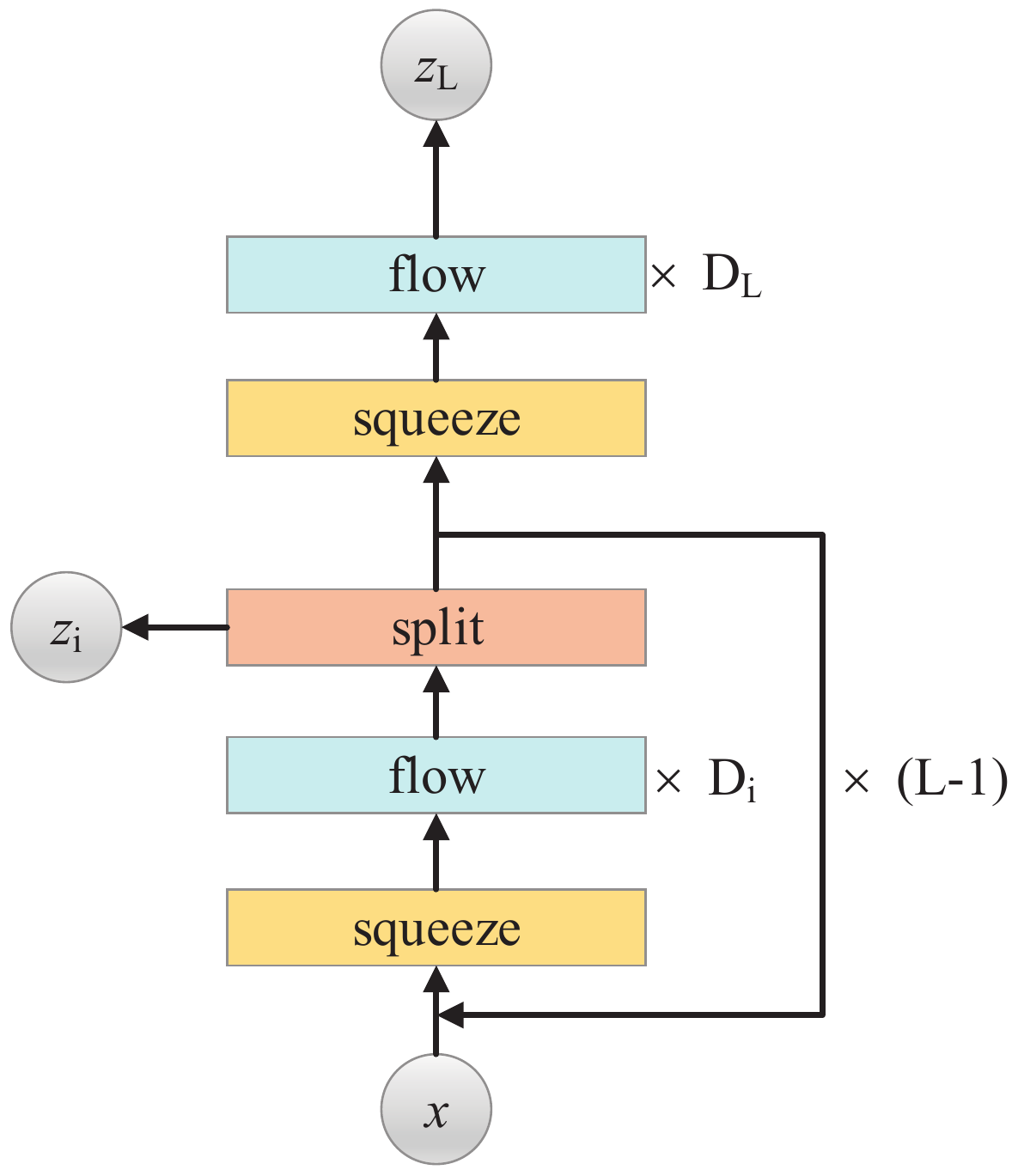}
		\end{minipage}}
		\caption{Overview of the model architecture. Left describes each step of flow, which consists of an actnorm layer that normalizes all activations independently, followed by a matrix exponential invertible $ 1\times 1 $ convolution, followed by a matrix exponential coupling layer. Right shows the multiscale architecture. The squeeze operation rearranges the dimensions by reducing the spatial dimensions by a half and increasing the channel number by four. The split operation splits the dimensions into two parts along channel and outputs a part of dimensions. The architecture has $ L $ levels and $ D_{i} $ flows for one level.}
		\label{fig:1}
	\end{center}
	\vskip -0.2in
\end{figure}

\section{Matrix Exponential Flows}

We utilize matrix exponential to propose a new flow called matrix exponential flows (MEF). In Section \ref{section:4.1}, we combine matrix exponential with coupling layers to present our matrix exponential coupling layers. In Section \ref{section:4.2}, we provide matrix exponential invertible $ 1\times 1 $ convolutions which are stable during training. Figure \ref{fig:1} illustrates a detailed overview of the architecture.

\subsection{Coupling layers}
\label{section:4.1}

\citet{dinh2016density} proposed affine coupling layers that split the $ n $ dimensional input $ \bm{x} $ into two parts $ (\bm{x}_{1:d},\bm{x}_{d+1,n}) $, the output $ \bm{y} $ of affine coupling layers follows the equations
\begin{equation}
\begin{aligned}
\bm{y}_{1:d}&=\bm{x}_{1:d}\\
\bm{y}_{d+1:n}&=\exp(\bm{s}(\bm{x}_{1:d}))\odot \bm{x}_{d+1:n}+\bm{b}(\bm{x}_{1:d})
\end{aligned}
\label{formula:9}
\end{equation}
where $ s $ and $ t $ stand for scale and bias, are functions from $ R^{d}\rightarrow R^{n-d} $, $ \exp(\bm{s}(\bm{x}_{1:d})) $ is the element-wise exponential function of $ \bm{s}(\bm{x}_{1:d}) $, and $ \odot $ is the Hadamard product. The first part remains unchanged and the second part is mapped with an element-wise exponential transformation, parametrized by the first part. Note that $ \bm{y}_{d+i} $ is only the function of $ \bm{x}_{1:d} $ and $ \bm{x}_{d+i} $, not the function of $ \bm{x}_{d+j}$, where $1\leq j \leq n-d,j\neq i $. Rewrite the affine coupling layers as
\begin{equation}
\left(
\begin{array}{c}
\bm{y}_{d+1} \\
\vdots\\
\bm{y}_{n} \\
\end{array}
\right)
=diag(\exp(\bm{s}(\bm{x}_{1:d}))
\left(
\begin{array}{c}
\bm{x}_{d+1} \\
\vdots\\
\bm{x}_{n} \\
\end{array}
\right)
+\bm{b}(\bm{x}_{1:d})
\label{formula:10}
\end{equation}
where $ diag(\exp(\bm{s}(\bm{x}_{1:d}))) $ is the diagonal matrix whose diagonal elements correspond to the vector $ \exp(\bm{s}(\bm{x}_{1:d})) $. As a diagonal matrix is less expressive, we replace the $ diag(exp(\bm{s}(\bm{x}_{1:d}))) $ by matrix exponential $ \bm{e}^{\bm{S}(\bm{x}_{1:d})} $, thus
\begin{equation}
\left(
\begin{array}{c}
\bm{y}_{d+1} \\
\vdots\\
\bm{y}_{n} \\
\end{array}
\right)
=\bm{e}^{\bm{S}(\bm{x}_{1:d})}
\left(
\begin{array}{c}
\bm{x}_{d+1} \\
\vdots\\
\bm{x}_{n} \\
\end{array}
\right)
+\bm{b}(\bm{x}_{1:d})
\label{formula:11}
\end{equation}
where $ \bm{e}^{\bm{S}(\bm{x}_{1:d})} $ is the matrix exponential of $ \bm{S}(\bm{x}_{1:d})\in R^{n-d\times n-d} $, each element of $ \bm{S}(\bm{x}_{1:d}) $ is a function of $ \bm{x}_{1:d} $. The first part is still unchanged. This form of layers is more expressive than former affine coupling layers. If $ \bm{S}(\bm{x}_{1:d})=diag(\bm{s}(\bm{x}_{1:d})) $, then $ \bm{e}^{\bm{S}(\bm{x}_{1:d})}=diag(\exp(\bm{s}(\bm{x}_{1:d}))) $, thus our coupling layers turn into affine coupling layers when the matrix $ \bm{S}(\bm{x}_{1:d}) $ is a diagonal matrix. Affine coupling layers are a special kind of our coupling layers. Matrix exponential of a $ 1\times 1 $ matrix is equal to exponential function, thus our coupling layers can be seen as multivariate affine coupling layers. 
The Jacobian of our coupling layers is 
\begin{equation}
\frac{\partial{\bm{y}}}{\partial{\bm{x}}}=
\left(
\begin{array}{cc}
\bm{I}_{d} & 0\\
\frac{\partial{\bm{y}_{d+1:n}}}{\partial{\bm{x}_{1:d}}} &  \bm{e}^{\bm{S}(\bm{x}_{1:d})}\\
\end{array}
\right)
\label{formula:12}
\end{equation}
The Jacobian is a block triangular matrix. Compared with affine coupling layers whose Jacobian is a triangular matrix, our coupling layers extend the Jacobian from a triangular matrix to a block triangular matrix. Its log-determinant is $ \log\det( \bm{e}^{\bm{S}(\bm{x}_{1:d})})=\trace(\bm{S}(\bm{x}_{1:d})) $, which can be computed fast. The inverse function of our layers is:
\begin{equation}
\begin{aligned}
\bm{x}_{1:d}&=\bm{y}_{1:d}\\
\bm{x}_{d+1:n}&=\bm{e}^{-\bm{S}(\bm{y}_{1:d})}(\bm{y}_{d+1:n}-\bm{b}(\bm{y}_{1:d}))
\end{aligned}
\label{formula:13}
\end{equation}

\begin{table*}[t]
	\caption{Density estimation performance on CIFAR-10 and ImageNet 32$ \times $32, ImageNet 64$ \times $64 datasets. Results are reported in bits/dim (negative $ \log_{2} $ likelihood). In brackets are models that use variational dequantization \cite{ho2019flow++}.
	}
	\label{table:2}
	\vskip 0.15in
	\begin{center}
		\begin{small}
			\begin{tabular}{llll}
				\toprule
				Model & CIFAR10 & ImageNet$32 \times 32$ & ImageNet$64 \times 64$ \\
				\hline
				RealNVP \cite{dinh2016density} & 3.49 & 4.28 & 3.98\\
				\hline
				Glow \cite{kingma2018glow} & 3.35 & 4.09 & 3.81\\
				\hline
				Emerging \cite{hoogeboom2019emerging} & 3.34 & 4.09 & 3.81\\
				\hline
				Flow++ \cite{ho2019flow++}& 3.29 (3.08) & ---  (3.86) & --- (3.69)\\
				\hline
				\textbf{MEF (Ours)} & 3.32 & 4.05 & 3.73\\
				\bottomrule
			\end{tabular}
		\end{small}
	\end{center}
	\vskip -0.1in
\end{table*}

\begin{table*}[t]
	\caption{Comparison of the number of parameters of Glow, Emerging, Flow++ and MEF}
	\label{table:3}
	\vskip 0.15in
	\begin{center}
		\begin{small}
			\begin{tabular}{llll}
				\toprule
				Model & CIFAR10 & ImageNet$32 \times 32$ & ImageNet$64 \times 64$ \\
				\hline
				Glow \cite{kingma2018glow} & 44.0M & 66.1M & 111.1M\\
				\hline
				Emerging \cite{hoogeboom2019emerging} & 44.7M & 67.1M & 67.1M\\
				\hline
				Flow++ \cite{ho2019flow++} & 31.4M & 169.0M & 73.5M\\
				\hline
				\textbf{MEF (Ours)} & 37.7M & 37.7M & 46.6M\\
				\bottomrule
			\end{tabular}
		\end{small}
	\end{center}
	\vskip -0.1in
\end{table*}

For 2D image data $ \bm{x} $ with shape $ h\times w\times 2c $, where $ h, w, 2c $ denote the height, width and number of channels, split $ \bm{x} $ along channel into two parts $ (\bm{x}^{1}, \bm{x}^{2}) $. The corresponding output is $ (\bm{y}^{1}, \bm{y}^{2}) $. Eq. (\ref{formula:11}) shows that each element of $ \bm{S}(\bm{x}^{1}) $ is a function of $ \bm{x}^{1} $. So the output layer of $ \bm{S}(\bm{x}^{1}) $ has $ (h\cdot w \cdot c)^2 $ units, which leads to too many units in the output layer of $ \bm{S}(\bm{x}^{1}) $. In order to reduce the number of units, we propose a location-dependent type of coupling layers. The output $ \bm{y}_{i,j,k}^{2} $ is not the function of all elements of $ \bm{x}^{2} $, but the function of $ \bm{x}_{i,j,l}^{2} $ with the same height and width index, where $ 1\leq l\leq c $. Our coupling layers turn into 
\begin{equation}
\begin{aligned}
\bm{y}^{1}&=\bm{x}^{1}\\
\forall i,j, \bm{y}^{2}_{i,j,:}&=\bm{e}^{\bm{S}(\bm{x}^{1})_{i,j,:,:}}\bm{x}^{2}_{i,j,:}+\bm{b}(\bm{x}^{1})_{i,j,:}
\end{aligned}
\label{formula:14}
\end{equation}
where $ \bm{S}(\bm{x}^{1})\in R^{h\times w\times c\times c} $,$ \bm{b}(\bm{x}^{1})\in R^{h\times w\times c} $. So the output layer of $ \bm{S}(\bm{x}^{1}) $ only has $ h\cdot w\cdot c\cdot c $ units. For very large $ c $, the output layer may still have too many units. Thus for very large $ c $, let the output layer of $ \bm{S}(\bm{x}_{1}) $ has $ h\times w\times c\times 2t $ units, where $ t  $ can be chosen such that $ t\ll c $. 
Split the output layer into two parts: $ \bm{A}_{1} $ with shape $ h\times w\times c\times t $ and $ \bm{A}_{2} $ with shape $ h\times w\times t\times c $. Our coupling layers follow
\begin{equation}
\begin{aligned}
\bm{y}^{1}&=\bm{x}^{1}\\
\forall i,j, \bm{y}^{2}_{i,j,:}&=\bm{e}^{\bm{A_{1}}(\bm{x}^{1})_{i,j,:,:}{\bm{A_{2}}(\bm{x}^{1})_{i,j,:,:}}}\bm{x}^{2}_{i,j,:}+\bm{b}(\bm{x}^{1})_{i,j,:}
\end{aligned}
\label{formula:15}
\end{equation}
And use Eq. (\ref{formula:7}) to compute matrix exponential. This makes the model scalable, we can select a proper $ t $ to balance the model complexity and computation. Since $ \bm{e}^{\bm{0}}=\bm{I}$, we initialize the output layer of $ \bm{S}(\bm{x}_{1}) $ with zeros such that each coupling layer initially performs an identity function, this helps training deep networks and reduces the computation of matrix exponential.

\subsection{Invertible 1$ \times $ 1 convolutions}
\label{section:4.2}

Standard 1$ \times $ 1 convolutions are flexible since the weight matrix $ \bm{W} $ can become any matrix in $ M_{n}(R) $. But they may be numerically unstable during training when the weight matrix is singular. \citet{kingma2018glow} proposed to learn a $ \bm{PLU} $ decomposition and constrained the diagonal element of $ \bm{U} $ non-zero, which makes the convolutions more stable, but their flexibility is limited. In order to solve the stability issues and retain the flexibility of the convolutions, we propose to replace the weight matrix $ \bm{W} $ by the matrix exponential of $ \bm{W} $, the convolutions are implemented as :
\begin{equation}
\forall i,j: \bm{y}_{i,j,:}=\bm{e}^{\bm{W}}\bm{x}_{i,j,:} 
\label{formula:16}
\end{equation}
In Section \ref{section:3.1}, we demonstrate that $ \bm{e}^{\bm{W}}\in GL_{n}(R)^{+} $, which guarantees the determinant of $ \bm{e}^{\bm{W}} $positive. So our matrix exponential convolutions are stable. The log-determinant of Jacobian is $ h\cdot w \cdot \trace(\bm{W}) $, where $ h,w $ are height and width.
The inverse function is:
\begin{equation}
\forall i,j: \bm{x}_{i,j,:}=\bm{e}^{\bm{-\bm{W}}}\bm{y}_{i,j,:} 
\label{formula:17}
\end{equation}
Suppose $ \bm{W} $ is a skew-symmetric matrix, then
\begin{equation}
(\bm{e}^{\bm{W}})(e^{\bm{W}})^T=\bm{e}^{\bm{W}+\bm{W}^{T}}=\bm{e}^{\bm{0}}=\bm{I}
\label{formula:18}
\end{equation}
thus matrix exponential of a skew-symmetric matrix is an orthogonal matrix. And the determinant of $ \bm{e}^{\bm{W}} $ is positive, so $ \bm{e}^{\bm{W}} $ is a rotation matrix. All $ n \times n $ rotation matrices form a special orthogonal group. Special orthogonal group is in the image of matrix exponential \cite{hall2015lie}. We initialize $ \bm{W} $ as a skew-symmetric matrix such that $ \bm{e}^{\bm{W}} $ is a rotation matrix.

%------------------------------------------------------------------------
\section{Related Work}
This work mainly builds upon the ideas proposed in \cite{dinh2016density,kingma2018glow}. Generative flows models can roughly be divided into two categories according to the Jacobian. One is the models whose Jacobian is a triangular matrix, which are based on coupling layers proposed in \cite{dinh2014nice,dinh2016density} or autoregressive flows proposed in \cite{kingma2016improved,papamakarios2017masked}.  \citet{ho2019flow++,hoogeboom2019emerging,durkan2019neural} extended the models with more expressive invertible functions. The other is the models with free-form Jacobian. \citet{behrmann2019invertible} proposed invertible residual networks and utilized it for density estimation. \citet{chen2019residual} further improved the model with a unbiased estimate of the log density. \citet{grathwohl2018ffjord} proposed a continuous-time generative flow with unbiased density estimation.

%------------------------------------------------------------------------

\section{Experiments}

In this section, we run several experiments to demonstrate the performance of our model. In Section \ref{section:6.1}, we compare the performance on density estimation with other generative flows models. In Section \ref{section:6.2}, we study the training stability of generative flows models. In Section \ref{section:6.3}, we compare three $ 1 \times 1 $ convolutions. In Section \ref{section:6.4}, we analyze the computation of matrix exponential. In Section \ref{section:6.5} we show samples from our trained models.

\subsection{Density Estimation}
\label{section:6.1}

We evaluate our MEF model on CIFAR10 \cite{krizhevsky2009learning}, ImageNet32 and ImageNet64 \cite{van2016pixel} datasets and compare log-likelihood with other generative flows models. See Figure \ref{fig:1} for a detailed overview of our architecture. We use a level $ L=3 $ and depth $ D_{1}=8, D_{2}=4, D_{3}=2 $. Each coupling layer is composed of 8 residual blocks \cite{he2016deep} for CIFAR10 and ImageNet32 datasets and 10 residual blocks for ImageNet64 dataset. Each residual block has three convolution layers, where the first layer and the last layer are $ 3\times 3 $ convolution layers, the center layer is $ 1\times 1 $ convolution layer, all with 128 channels. The activation function is ELU \cite{clevert2015fast}. The optimization method is Adamax \cite{kingma2014adam}. All models are trained for 50 epochs with batch size 64. Table \ref{table:2} shows MEF achieves great performance on the negative log-likelihood scores in bits/dim. Our model performs better than Glow \cite{kingma2018glow} and Emerging \cite{hoogeboom2019emerging}, only worse than Flow++\cite {ho2019flow++} with variational dequantization. Table \ref{table:3} shows the comparison of the number of parameters of Glow, Emerging, Flow++ and MEF. Our model uses a relatively small number of parameters on ImageNet datasets.

\subsection{Training Stability}
\label{section:6.2}

\begin{figure}[t]
	\vskip 0.2in
	\begin{center}
		\centerline{\includegraphics[width=1.0\linewidth]{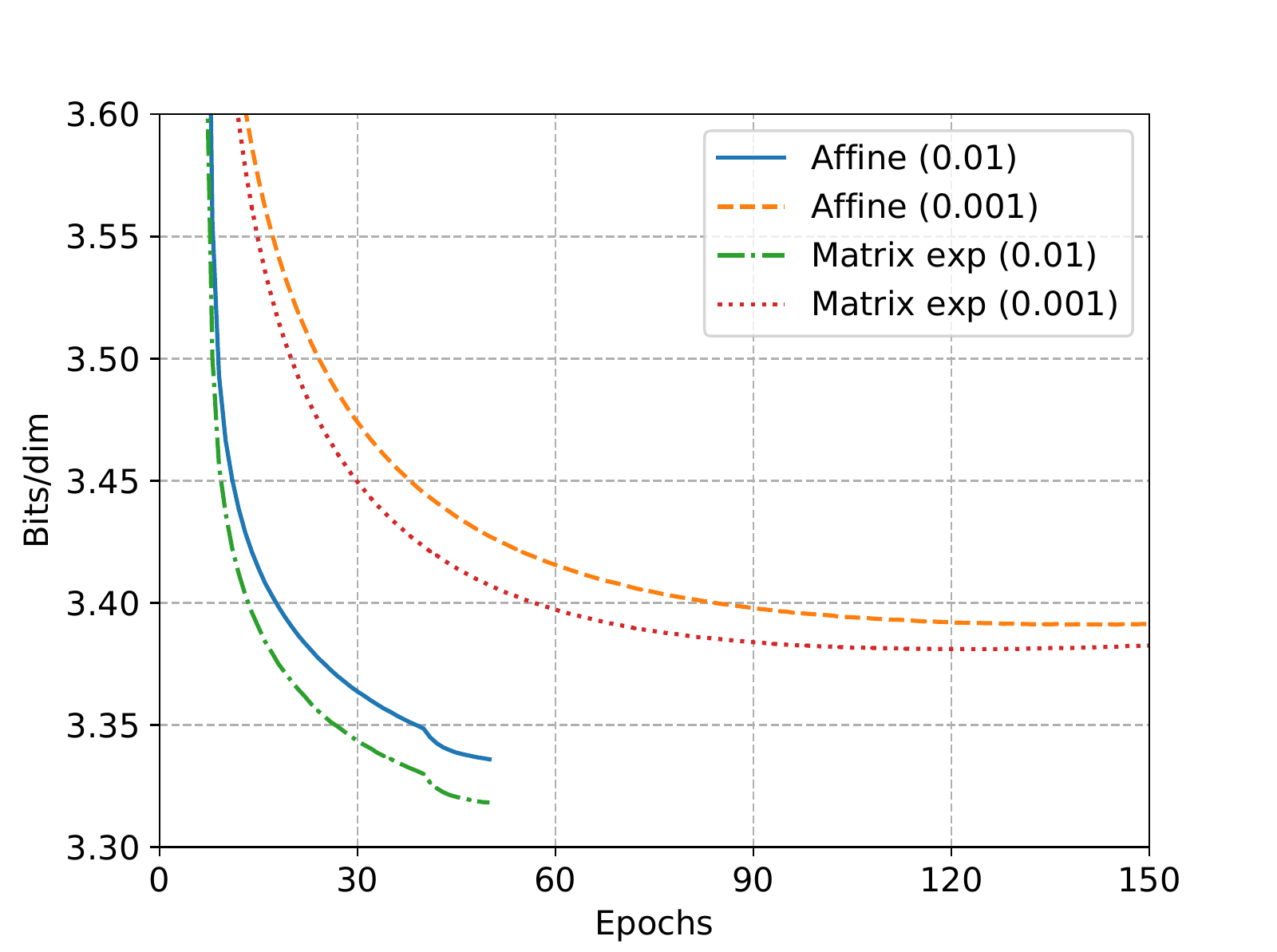}}
		\caption{Bits per dimension curve on CIFAR10 test set with different coupling layers and learning rate.}
		\label{fig:2}
	\end{center}
	\vskip -0.2in
\end{figure}

Training generative flows models requires high computing infrastructure due to large computation, especially for large-scale datasets. One reason is that training generative flows models often take a long time to converge. Glow \cite{kingma2018glow} was trained for 1800 epochs and Flow++ \cite{ho2019flow++} had not fully converged after 400 epochs on CIFAR10 dataset. One reason is that training generative flows models is not stable, which prevents us from being able to use a larger learning rate. In Section \ref{section:4.2}, we explain why standard $1\times 1 $ convolutions may collapse during training. We  replace the standard convolutions by our matrix exponential convolutions, which makes the convolutions stable. In our experiments, we find that training generative flows models often diverge when using coupling layers. The reason is that the output of model may be pretty large when using coupling layers. The log-likelihood in Eq. (\ref{formula:2}) is composed of two terms. $ p_{Z}(z) $ often chooses Gaussian distribution. When the output is pretty large, $ p_{Z}(z) $ tends to zero, then the log-likelihood tends to infinity. Coupling layers can be written as:
\begin{equation}
\begin{aligned}
\bm{y_{1}}&=\bm{x_{1}}\\
\bm{y_{2}}&=g(\bm{x_{2}};c(\bm{x_{1}}))
\end{aligned}
\label{formula:19}
\end{equation}
where $ c $ is a function of $ \bm{x_{1}} $ and $ g $ is an invertible function with respect to $ \bm{x_{2}} $. \citet{dinh2016density} used hyperbolic function to make the output of $ c $ bounded. We further extend the idea to control the value of the output of $ c $ to prevent divergence during training. 
For matrix exponential coupling layers, we modify Eq.(\ref{formula:11}) to:
\begin{equation}
\bm{y_{2}}=\bm{e}^{(u_{1}\tanh(u_{2} \bm{S}+v_{2})+v_{1})}\bm{x_{2}}+\bm{b}
\label{formula:20}
\end{equation}
where $ u_{1}, v_{1}, u_{2}, v_{2} $ are scalar parameter, which can be learned during training, and $ \tanh(\cdot) $ is hyperbolic function. Initialize $ v_{1}=0, v_{2}=0 $. We first initialize $ u_{1}=1 $ and $ u_{2}=1 $. If the model diverges, then we initialize $ u_{1} $ and $ u_{2} $ with smaller values. We repeat this operation until convergence.
Using this form of coupling layers can make training more stable and allows us to use a larger learning rate.
Affine coupling layers have the similar form:
\begin{equation}
\bm{y_{2}}=\exp(u_{1}\tanh(u_{2} \bm{s}+v_{2})+v_{1})\odot\bm{x_{2}}+\bm{b}
\label{formula:21}
\end{equation}
We run models on CIFAR10 dataset with learning rate 0.01 and 0.001 to compare the convergence speed. Models with learning rate 0.01 are trained for 50 epochs, and models with learning rate 0.001 are trained for 150 epochs. We also compare our matrix exponential coupling layers with affine coupling layers. Figure \ref{fig:2} and table \ref{table:4} show the results. Using a learning rate 0.01 achieves better performance and converges faster than using a learning rate 0.001. The results also show that matrix exponential coupling layers perform better than affine coupling layers.
\begin{table}[t]
	\caption{Comparison of models with different coupling layers and learning rate. Performance is measured in bits per dimension. In brackets are the learning rate. Results are obtained by running 3 times
		with different random seeds, $ \pm $ reports standard deviation.}
	\label{table:4}
	\vskip 0.15in
	\begin{center}
		\begin{small}
			\begin{tabular}{ll}
				\toprule
				Model & CIFAR10  \\
				\hline
				Affine (0.01) & 3.336$ \pm $ 0.002 \\
				\hline
				Affine (0.001) & 3.391$ \pm $ 0.003 \\
				\hline
				Matrix exp (0.01) & 3.324$ \pm $ 0.004 \\
				\hline
				Matrix exp (0.001) & 3.381$ \pm $ 0.008 \\
				\bottomrule
			\end{tabular}
		\end{small}
	\end{center}
	\vskip -0.1in
\end{table}

\begin{figure*}[t]
	\vskip 0.2in
	\begin{center}
		\includegraphics[width=\linewidth]{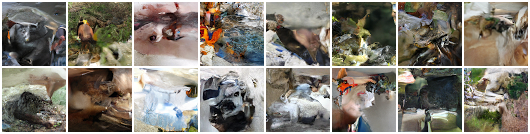}
		\caption{Samples from our trained ImageNet64 model}
		\label{fig:3}
	\end{center}
	\vskip -0.2in
\end{figure*}
\begin{figure}[t]
	\vskip 0.2in
	\begin{center}
		\includegraphics[width=\linewidth]{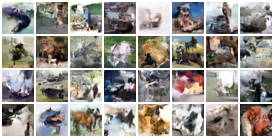}
		\caption{Samples from our trained CIFAR10 model}
		\label{fig:4}
	\end{center}
	\vskip -0.2in
\end{figure}
\begin{figure}[t]
	\vskip 0.2in
	\begin{center}
		\includegraphics[width=\linewidth]{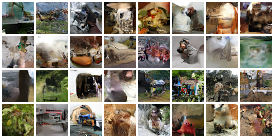}
		\caption{Samples from our trained ImageNet32 model}
		\label{fig:5}
	\end{center}
	\vskip -0.2in
\end{figure}

\subsection{$ 1\times 1 $ Convolutions}
\label{section:6.3}

We run models on CIFAR10 dataset to compare the performance of standard $ 1\times 1 $ convolutions, $ \bm{PLU} $ decomposition $ 1 \times 1 $ convolutions and  matrix exponential $ 1 \times 1 $ convolutions. All models have the same parameter settings expect the $ 1\times 1 $ convolutions. We also record the running time per epoch to compare the computation of convolutions. All models are trained on one TITAN Xp GPU. Table \ref{table:5} shows the result. Our matrix exponential $ 1\times1 $ convolutions achieve nearly same performance on density estimation as standard convolutions and have nearly the same computation compared with $ \bm{PLU} $ decomposition $ 1 \times 1 $ convolutions. 

\begin{table}[h]
	\caption{Comparison of standard, $ \bm{PLU} $ decomposition and matrix exponential convolutions. Performance is measured in bits per dimension. Computation is measured in running time per epoch. Results are obtained by running 3 times
		with different random seeds, $ \pm $ reports standard deviation.}
	\label{table:5}
	\vskip 0.15in
	\begin{center}
		\begin{small}
			\begin{tabular}{lll}
				\toprule
				Convolutions & CIFAR10 & Time \\
				\hline
				Standard & 3.324$ \pm $ 0.001  & 844.3$ \pm $ 2.6s \\
				\hline
				Decomposition & 3.330$ \pm $ 0.007 & 668.3$ \pm $ 19.6s  \\
				\hline
				Matrix exp & 3.324$ \pm $ 0.004 & 669.5$ \pm $ 10.1s \\
				\bottomrule
			\end{tabular}
		\end{small}
	\end{center}
	\vskip -0.1in
\end{table}

\subsection{Truncate Matrix Exponential}
\label{section:6.4}

Matrix exponential is an infinite matrix series. We need to truncate it at a finite term to approximate it. We use  Algorithm \ref{alg:1} to approximate it, which costs about $ (s+k-1)n^{3} $ FLOPs. In this section, we present the coefficient $ m=s+k-1 $ during our training. We set the tolerable error of Algorithm \ref{alg:1} $ \epsilon=10^{-8} $. We count 1 million times of coefficient $ m $ when computing matrix exponential. In Table \ref{table:6}, we show the mean, standard deviation, maximum and minimum of coefficient $ m $. The coefficient $ m $ is no more than 11 and is about 9 in average. Experiments show that matrix exponential can converge fast.
\begin{table}[h]
	\caption{Mean, standard deviation, maximum and minimum of the coefficient $ m $.}
	\label{table:6}
	\vskip 0.15in
	\begin{center}
		\begin{small}
			\begin{tabular}{lllll}
				\toprule
				& Mean & Std & Max & Min  \\
				\hline
				Coefficient $ m $ & 9.28 & 0.94 & 11 & 2\\
				\bottomrule
			\end{tabular}
		\end{small}
	\end{center}
	\vskip -0.1in
\end{table}

\subsection{Samples}
\label{section:6.5}

We show the samples from our trained models on CIFAR10, ImageNet32 and ImageNet64 datasets in Figure \ref{fig:3} to     \ref{fig:5}. Our CIFAR10 model takes 1.67 seconds to generate a batch of 64 samples on one NVIDIA 1080 Ti GPU.

%------------------------------------------------------------------------

\section{Conclusion}

In this paper, we propose a new type of generative flows, called matrix exponential flows, which utilizes the properties of matrix exponential. We incorporate matrix exponential into neural networks and combine it with generative flows. We propose matrix exponential coupling layers which are a generalization of affine coupling layers. In order to solve the stability problem, we propose matrix exponential $ 1\times 1 $ convolutions and improve the coupling layers. Our model significantly speeds up the training process. Based on matrix exponential, we hope that more layers can be proposed or incorporate it into other layers.

\section*{Acknowledgements}

We thank the reviewers for their insightful comments. This work is supported by the National Natural Science Foundation of China (61672482) and Zhejiang Lab (NO. 2019NB0AB03).


\bibliography{example_paper}
\bibliographystyle{icml2020}

\end{document}